\documentclass{article}
\usepackage{spconf,amsmath,epsfig}
\usepackage{float}
\usepackage{graphicx}
\usepackage{subfigure}
\usepackage{multirow}

\begin{document}\sloppy

\def\x{{\mathbf x}}
\def\L{{\cal L}}

\title{Light Weight Color Image Warping with Inter-Channel Information}
%
\name{Chuangye Zhang,
	YanNiu,
	Tieru Wu,
	Ximing Li}
\address{}

\maketitle

\begin{abstract}
Image warping is a necessary step in many multimedia applications such as texture mapping, image-based rendering, panorama stitching, image resizing and optical flow computation etc. Traditionally, color image warping interpolation is performed in each color channel independently. In this paper, we show that the warping quality can be significantly enhanced by exploiting the cross-channel correlation. We design a warping scheme that integrates intra-channel interpolation with cross-channel variation at very low computational cost, which is required for interactive multimedia applications on mobile devices. The effectiveness and efficiency of our method are validated by extensive experiments. 
\end{abstract}

\section{Introduction}
\label{sec:intro}
Image warping is fundamental for a variety of multimedia applications such as image resizing (e.g, \cite{ledig2017photo}), texture mapping (e.g., \cite{woo1999opengl}), image-based rendering (e.g., \cite{mcmillan2009image}), panorama stitching (e.g., \cite{he2013rectangling}), stereo reconstruction (e.g., \cite{bleyer2009stereo}), optical flow computation (e.g., \cite{bouguet2001pyramidal}), to name but a few. Image warping can be briefly described as transforming a source image $I_{1}$ into a target image $I_{2}$ under a geometric mapping $\mathbf{H}$, which can be linear, affine, perspective, non-parametric etc.\cite{glasbey1998review}. Particularly, assuming $I_{2}$ has $M$ rows and $N$ columns, each pixel $(i,j)$ in the target image lattice $[1 \quad M]\times[1 \quad N]$ has a correspondence image point $[x,y]=\mathbf{H}^{-1}([i,j])$ in the source image, from which the intensity value $I_{1}[x,y]$ is assigned to $I_{2}[i,j]$. The central problem of image warping is to estimate $I_{1}[x,y]$ from pixels in the vicinity of $[x,y]$, which are generally not on the integer grid, and hence the true intensity $I_{1}[x,y]$ is not immediately available. 

The most popular estimation techniques for image warping in real practice are nearest neighbour, bilinear and bicubic (spline) interpolation schemes \cite{zitova2003image}, which do not adapt well to irregular image regions. Therefore, academic research effort has been mainly dedicated to edge guided interpolation, such as anisotropic filtering (e.g., \cite{battiato2002locally}, \cite{li2001new}) and variational regularization (e.g., \cite{jiang2002new}). Particularly on the sub-topic of Signal Image Super-resolution, i.e., in the special case that $\mathbf{H}$ degenerates to a scaling function, mathematical models such as compressive sensing (e.g., \cite{yang2010image}), dictionary learning (e.g., \cite{yang2012coupled}) and Convolutional Neural Networks (CNN, e.g., \cite{ledig2017photo}) have also been intensively investigated.  

Although numerous image warping schemes have been proposed, they are mostly designed for monochrome images. This means that on warping color images, these methods should be applied to each channel independently (e.g., \cite{su2004image}) or to the luminance channel only (e.g.\cite{dong2014learning}), without exploiting the cross-channel correlation. The only exception is end-to-end CNNs that take RGB images as input. In this situation, the color correlation is captured during the training process, but the learned correlation is CNN specific rather than being general. In contrast, cross-channel correlation is thoroughly studied for Image Demosaicking, which recovers the full RGB images from subsampled color channels, with only one primary color component available at each pixel \cite{szeliski2010}. Although each channel can be recovered by merely using the intra-channel interpolation, it is commonly known that inter-channel correlation information can significantly improve the demosaicking accuracy. 

In this paper, we integrate intra-channel image warping with cross-channel correlation, and show that our approach significantly improve the warping accuracy at very low cost. The proposed integration scheme is inspired by the Malvar-He-Cutler High Quality Linear Interpolation (HQLI) demosaicking method \cite{malvar2004high}, which is probably the fastest demosaicking algorithm in the literature (except the Nearest Neighbour and Bilinear interpolation), but its trade-off between accuracy and speed is superior to many sophisticated algorithms \cite{Niu2018TIP}. We first examine the potential of cross-channel correlation for image warping, by applying HQLI to postprocess color images upsampled with intra-channel interpolation. We then generalize this combination to formulate general color image warping. The performance of our method is assessed in the scenario of image super-resolution, as it is the most active sub-area of image warping. Despite its very low cost, the proposed algorithm achieves comparable accuracy to state of the art algorithms at about $40$ times faster speed. Compared to CNN-based methods, our model has the flexibility of resizing the image to arbitrary size. These features are desirable for real time interactive applications on multimedia devices that have limited computation power (e.g., mobile smart phones).
\section{Inter-Channel Correlation Gain}
\label{sec:motivation}
Image warping is generally implemented by backward mapping. Given $\mathbf{H}$ the geometric mapping from the source image $I_{1}$ to the target image $I_{2}$, and $\mathbf{H}^{-1}$ be its inverse mapping, $\mathbf{H}^{-1}$ maps the coordinate frame of $I_{2}$ back to the coordinate frame of $I_{1}$. Image $I_{2}$ is generated in such a way that, for each pixel $(i,j)$ of $I_{2}$, the intensity value $I_{2}[i,j]$ is copied from $I_{1}[x,y]$, where
\begin{align}
\left[
\begin{array}{c}
	x\\y
\end{array}
\right] = \mathbf{H}^{-1}
\left[
\begin{array}{c}
	i\\j
\end{array}
\right].
\end{align} 
As $x$ and $y$ are generally not integers, $I_{1}[x,y]$ has to be estimated from the surrounding pixels. Let $m = \left\lfloor x \right\rfloor$ and $n = \left\lfloor y \right\rfloor$ be the floor integer parts of $x$ and $y$, and define $s=x-m$, $t=y-n$. For example, the bilinear warping scheme estimates $I_{2}[i,j]$ by 
\begin{align}
	\begin{split}
	\hat{I}_{1}[x,y] & = (1-s)(1-t){I}_{1}[m,n]+st{I}_{1}[m+1,n+1]\\
	                 & + s(1-t){I}_{1}[m+1,n]+(1-s)t{I}_{1}[m,n+1].
	\end{split}
	\label{eq:linearWarp}
\end{align}
Today, multimedia images are usually chromatic rather than being grayscale. Therefore, $I_{1}$ and $I_{2}$ generally have RGB three channels. Typically, bilinear warping for color images is to replace $I_{1}$ in Eq.\ref{eq:linearWarp} with its color channels $R_{1}$, $G_{1}$ and $B_{1}$, to estimate $R_{2}[i,j]$, $G_{2}[i,j]$ and $B_{2}[i,j]$, without cross-channel interaction.

\begin{figure}[h]
\begin{center}
	\subfigure[RGGB]{
	\label{fig:Bayer:a} 
	\includegraphics[width=0.25\linewidth]{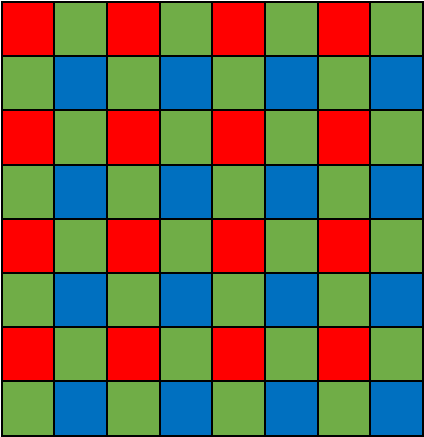}}
    \hspace{0.5in}
    \subfigure[BGGR]{
    \label{fig:Bayer:b} 
    \includegraphics[width=0.25\linewidth]{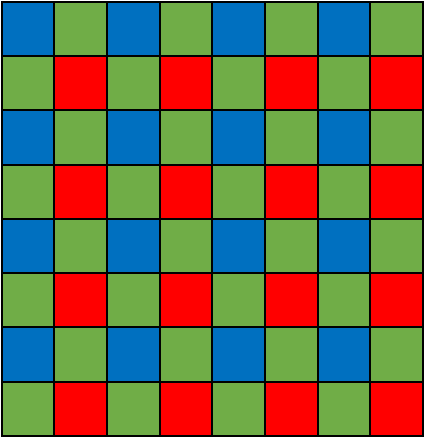}}
    \hspace{0.5in}
	\subfigure[GRBG]{
	\label{fig:Bayer:c} 
	\includegraphics[width=0.25\linewidth]{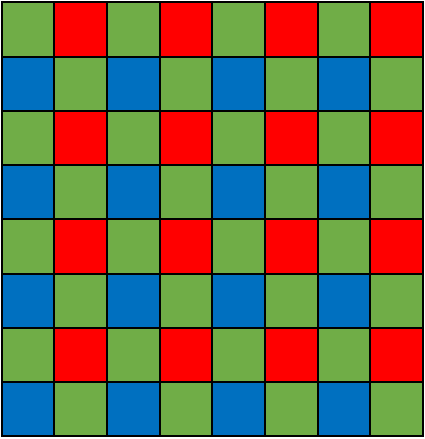}}
    \hspace{0.5in}
    \subfigure[GBRG]{
	\label{fig:Bayer:d} 
	\includegraphics[width=0.25\linewidth]{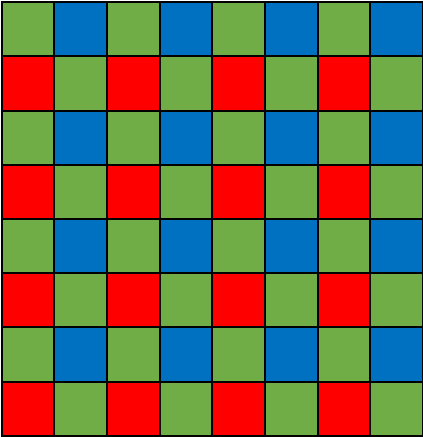}}
\hspace{0.5in}
\end{center}
  \caption{An illustration of the Bayer pattern for CFA.}
  \label{fig:Bayer}
\end{figure}

The importance of cross-channel correlation is well investigated for the problem of Image Demosaicking, entailed in current consumer-level digital cameras, which commonly use Color Filter Arrays (CFA) to sample one of the RGB components for each pixel. Fig.\ref{fig:Bayer} shows an example of the Bayer pattern CFA, adopted for most digital cameras. 

Define $\mathcal{R}$, $\mathcal{G}$ and $\mathcal{B}$ to be the sets that collect the pixels where the original red, green and blue values are captured respectively. Malvar-He-Cutler propose estimating the missing color values by a weighted combination of the local intra-channel average and the Laplacian of another channel \cite{malvar2004high}. For example, at a pixel $[m,n]\in \mathcal{R}$, the green value $G[m,n]$ is estimated by
\begin{equation}
G[m,n]\approx \bar{G} + \alpha \Delta R[m,n],
\label{eq:HQLI}
\end{equation}
where constant $\alpha$ is learned from training images, $\bar{G}$ stands for the average of $G[m,n-1]$, $G[m,n+1]$, $G[m-1,n]$ and $G[m+1,n]$, and 
\begin{align}
\begin{split}
\Delta R[m,n]  = R[m,n]-\frac{1}{4}&\left(R[m+2,n]+R[m-2,n]\right.\\
                   &\left.+R[m,n-2]+R[m,n+2]\right).\\
\end{split}
\label{eq:Laplacian}
\end{align}
The other missing color values are recovered in the same fashion. This demosaicking method, namely HQLI, adds second-order details (the Laplacian) of one channel to the average of another channel, thereby it performs image enhancement. 

To examine whether the inter-channel information supplements intra-channel recovery in image warping. We conduct a preliminary experiment using image super-resolution as an example application. We upsample test images by bilinear and bicubic interpolation, and then refine the upsampled image values by sequential HQLI demosaicking. In particular, we first apply the GRBG CFA pattern to filter the upsampled color image, and estimate the filtered-out color values by the remaining color values. Subsequently, we apply the RGGB CFA to the demosaicked full RGB image, and perform the second-round demosaicking. We then change the CFA pattern to BGGR, and repeat HQLI demosaicking for the final round. In this way, each color component of a pixel is refined by the intensity values in the local neighbourhood of the other two channels. Table\ref{tab:IndividualMcM} compares the reconstruction accuracy, measured by Peak Signal to Noise Ratio (PSNR), before and after demosaicking (i.e., the cross-channel information) on benchmark datasets Set5 \cite{bevilacqua2012low}, Set14 \cite{zeyde2010single}, BSD100 \cite{martin2001database} and Urban100 \cite{huang2015single}. It can be seen that the inter-channel correlation introduced by demosaicking notably improves the intra-channel interpolation, on all test datasets. Numerically, the improvement is upto $1.57$ dB and $1.09$ dB for bilinear and bicubic intra-channel interpolation respectively.
\begin{table}[!h]
  \centering
  	  \caption{The average PSNR before and after refining bilinear and bicubic upsampling by sequential HQLI demosaicking, tested on benchmark datasets.}
  	\label{tab:IndividualMcM}%
  	\vspace{1ex}

   \begin{tabular}{|c|c|c|c|c|}
			\hline
			{Dataset} & {Bilinear} & {\begin{tabular}{@{}c@{}}Bilinear\\+HQLI\end{tabular}} & {Bicubic} & {\begin{tabular}{@{}c@{}}Bicubic\\+HQLI\end{tabular}}\\
			\hline
			{Set5} & {28.86} & {30.43} & {28.64} & {29.73} \\
			{Set14} & {26.55} & {27.50} & {26.23} & {26.85} \\
			{BSD100} & {26.44} & {26.89} & {26.02} & {26.16} \\
		  {Urban100} & {23.57} & {24.12} & {23.17} & {23.45}\\
      \hline
    \end{tabular}%
\end{table}%

Note that to simulate the real world applications, here the low resolution images are obtained by Nearest Neighbour downsampling of the high resolution test images. This is different from the common SISR algorithm evaluation routine, which utilizes Bicubic downsampling. The Nearest Neighbour sampling is known to suffer severe aliasing and blurring, whereas Bicubic downsampling implicitly performs low pass filtering before sampling, thereby largely avoids such artifacts \cite{szeliski2010}. However, using Bicubic downsampling also implicitly assumes that the high resolution images are known, which is not the case for real applications. Therefore we only apply bicubic downsampling in the comparison experiments for fair conditions. Due to the coarse downsampling, bicubic warping has lower accuracy than bilinear warping, which is seemingly counter intuitive. 
\section{Cross-Channel Color Image Warping}
\label{sec:formulation}

\subsection{General Formulation}
Based on the analysis of the gain achieved by the inter-channel information, we formulate general color image warping by 
\begin{subequations} \label{eq:formula}
\begin{align}
G_{1}[x,y] &\approx& \tilde{G}_{1}[x,y]+\omega_{g,r}\Delta R_{1}[x,y]+\omega_{g,b}\Delta B_{1}[x,y] \label{eq:formula.1}\\
R_{1}[x,y] &\approx& \tilde{R}_{1}[x,y]+\omega_{r,g}\Delta G_{1}[x,y]+\omega_{r,b}\Delta B_{1}[x,y] \label{eq:formula.2}\\
B_{1}[x,y] &\approx& \tilde{B}_{1}[x,y]+\omega_{b,g}\Delta G_{1}[x,y]+\omega_{b,r}\Delta R_{1}[x,y] \label{eq:formula.3},
\end{align}
\end{subequations} 
where $\tilde{G}_{1}[x,y]$, $\tilde{R}_{1}[x,y]$ and $\tilde{B}_{1}[x,y]$ can be any intra-channel estimation of $G_{1}[x,y]$, $R_{1}[x,y]$ and $B_{1}[x,y]$; also, $\Delta G_{1}[x,y]$, $\Delta R_{1}[x,y]$ and $\Delta B_{1}[x,y]$ can be estimated by interpolating ${\Delta G_{1}[k,l]}_{[k,l]\in \Omega(m,n)}$, ${\Delta R_{1}[k,l]}_{[k,l]\in \Omega(m,n)}$ and ${\Delta B_{1}[k,l]}_{[k,l]\in \Omega(m,n)}$ respectively, where $\Omega[m,n]$ stands for the local neighbourhood of pixel $[m,n]$. The involved interpolation schemes can be either linear or non-linear, isotropic or anisotropic, depending on the particular requirements of the applications at hand. The weights $\omega_{(\cdot,\cdot)}$are to be learned from training images, as described in Section\ref{subsec: training}. 


\subsection{Parameter Training}
\label{subsec: training}
We illustrate the weight learning process using the Single Image Super Resolution (SISR) application as an example, for which the geometric mapping $\mathbf{H}$ and its inverse mapping $\mathbf{H}^{-1}$ are  
\begin{align}
\mathbf{H} = 
\left[\begin{array}{cc}
 S & 0 \\  0 & S
\end{array}\right]\quad
\mathbf{H}^{-1} = 
\left[\begin{array}{cc}
 \frac{1}{S} & 0 \\  0 & \frac{1}{S}
\end{array}\right], 
\end{align}
where $S$ in a real number larger than $1$. Note that $S$ is not restricted to be an integer. We downsample the groundtruth images as the source images $\mathbf{I}_{1}$ (i.e., the data) by bicubic interpolation, following the traditional experimental setting in the literature of SISR, as the learned weights are to be used for algorithm evaluation in comparison with state of the art. We use the groundtruth images (taken from benchmark training dataset BSD200 \cite{martin2001database}) as target images $\mathbf{I}_{2}$ (i.e., the labels). For each pixel $[i,j]$ in the an image $\mathbf{I}_{2}$, the RGB values of its correspondence $(x,y)$ in $\mathbf{I}_{1}$ thus are known in the training process. $\tilde{G}_{1}[x,y]$, $\tilde{R}_{1}[x,y]$ and $\tilde{B}_{1}[x,y]$, as well as $\Delta G_{1}[x,y]$, $\Delta R_{1}[x,y]$ and $\Delta B_{1}[x,y]$ are computed on the pixels in each source image by interpolation function $f$. Subsequently, Eq.\ref{eq:formula} can be rewritten as linear systems of the weights $\omega_{(\cdot,\cdot)}$. For example,
\begin{align}
\begin{split}
&\left[\begin{array}{cc}
\Delta R_{1}[x_1,y_1] & \Delta B_{1}[x_1,y_1]\\
\vdots & \vdots\\
\Delta R_{1}[x_K,y_K] & \Delta B_{1}[x_K,y_K]
\end{array}\right]
\left[\begin{array}{c}
\omega_{g,r}\\ \omega_{g,b}
\end{array}\right]  \\
& =\left[\begin{array}{c}
\tilde{G}_{1}[x_1,y_1]-G_{2}[i_1,j_1]\\
\vdots\\
\tilde{G}_{1}[x_K,y_K]-G_{2}[i_K,j_K]
\end{array}\right]
\end{split}
\label{eq:training}
\end{align} 
for Eq.\ref{eq:formula.1}. Here integer $K=10000$ is the total number of randomly selected training pixels, indexed by the subscripts of $x$, $y$, $i$ and $j$. Eq.\ref{eq:formula.2} and Eq.\ref{eq:formula.3} can be re-organized similarly. The Mean Square Error (MSE) solutions or Mean Absolute Error (MAE) solutions to the linear systems can be easily solved in closed-form. This yields the weights $\omega_{g,r}$, $\omega_{g,b}$,$\omega_{r,g}$,$\omega_{r,b}$,$\omega_{b,r}$ and $\omega_{b,g}$.

Seemingly, the learned weight values should be specific to the scaling factor $S$ and interpolation function $f$. However, we observe that weights learned for large $S$ actually generalizes well to small $S$. Therefore, we suggest conducting the training for a large scaling factor (e.g., $S=4$), and apply the learned weights directly to applications whose scaling factor is smaller.   

Table\ref{tab:weights} lists the weights we learned from Eq.\ref{eq:training} by linear regression for $S=4$, with the interpolation function $f$ being Bilinear, Bicubic and Lanczos (with $5\times 5$ footprint). 

\begin{table}[!h]
  \centering
  	
  \caption{The integration weights learned by linear regression for $S=4$ and various interpolation functions $f$.}
  \label{tab:weights}%
  \vspace{1ex}
   \begin{tabular}{|c|c|c|c|}
			\hline
		
			{} & {Bilinear} & {Bicubic} & {Lanczos}\\
			\hline
			{$\omega_{g,r}$} & {0.094} & {0.045} & {0.032}  \\
			\hline
			{$\omega_{g,b}$} & {0.119} & {0.064} & {0.041} \\
			\hline
			{$\omega_{r,g}$} & {0.195} & {0.096} & {0.058} \\
			\hline
		    {$\omega_{r,b}$} & {0.008} & {0.010} & {0.015} \\
			\hline
			{$\omega_{b,g}$} & {0.180} & {0.089} & {0.054} \\
           \hline
			{$\omega_{b,r}$} & {-0.003} & {0.003} & {0.008} \\
		  \hline
		 
    \end{tabular}%
	
\end{table}%

\section{Experimental Results}
\label{sec:experiments}
We conduct two sets of experiments to evaluate the proposed color image warping technique. The first experiment measures the SISR accuracy before and after incorporating the cross-channel terms to intra-channel warping, at various scaling factors and using different real time interpolation functions. The second experiments compares our technique with the state of the art SISR method Global Regression (GR) proposed by Timofte-De Smet-Van Gool in \cite{timofte2013anchored}, using the released Matlab source code. The GR method is the fastest SISR method proposed in recent years to the best of our knowledge, and hence is chosen for comparison in our experiment. It should be mentioned that, the weight GR is significantly higher than our linear isotropic scheme used for testing. 

For comparison in equal conditions, the low resolution test images $I_{1}$ are obtained by bicubicly downsampling the ground truth image $I_{2}$ at sampling factor $S$. That is, the low resolution images have $\left\lfloor \frac{M}{S} \right\rfloor$ rows and $\left\lfloor \frac{N}{S} \right\rfloor$ columns. Laplacians are computed at each pixel in each channel of the low resolution images. Both the Laplacian maps and low resolution images are upsampled by the same interpolation function $f$. In our experiments, $S$ is set to be 2, 3 or 4; and $f$ is set to be Bilinear, Bicubic or Lanczos (of kernel size $5 \times 5$).    

Evaluation is carried out on benchmark datasets Set5, Set14, BSD100 and Urban100. Beside the PSNR accuracy measure, we also report the Structure Similarity (SSIM) measure \cite{wang2004image} and running time (in seconds) of the SISR algorithms. All experiments are conducted on an Intel Core i7-7700 3.60GHz CPU with 8GB RAM, using Matlab.

\subsection{Improvement on Popular Warping Methods} 
We compare the performance of popular real time warping methods and the proposed method. Bilinear warping probably has the best trade-off between performance and speed in the literature, as pointed out by Zitov\'{a}-Flusser \cite{zitova2003image}. Hence it is recommended for video and animation applications, such as frame registration for motion estimation in OpenCV and texture mapping in OpenGL. Bicubic warping is another frequently adopted option, especially for still image processing. For example, image editing softwares Photoshop and GIMP employ bicubic interpolation for image resizing and perspective view rectification.

Table\ref{tab:improvement} presents the PSNR, SSIM and running time averaged over Set5 for various $S$ and $f$. In this table, the plain intra-channel warping is referred to as ``Independent'', and the proposed inter-channel color warping is referred to as ``Correlated''. In all situations, the proposed color warping scheme achieves evidential improvement, either in terms of PSNR or SSIM, in real time computation. The most remarkable quantitative improvement is obtained for bilinear interpolation, the PSNR values of which are increased by 1.58dB, 0.77dB and 0.98dB for 2, 3 and 4 times upscaling respectively. It can be observed that the improvement decreases with the complexity of the interpolation function. However, in the case of Lanczos interpolation (the most complex function here), the accuracy of our color warping scheme for $2\times$ upscaling is probably sufficient in many applications, and its computation is merely in several milliseconds.

\begin{table*}[!ht]
  \centering
  
  \caption{PSNR, SSIM and computation time of plain intra-channel warping and the proposed color warping, averaged over Set5, for various scaling factor $S$ and interpolation function $f$.}
  \label{tab:improvement}%
  \vspace{1ex}
  	\begin{tabular}{|c|c|c|c|c|c|c|c|}
  		       \hline
			
				\multirow{2}*{$f$}& \multirow{2}*{$S$}& \multicolumn{3}{|c|}{Independent} & \multicolumn{3}{|c|}{Correlated}  \\
				\cline{3-8}
				& & PSNR& SSIM & Time& PSNR& SSIM & Time\\
				\hline
				\multirow{3}*{Bilinear} &{$\times2$ }&30.42&	0.9566&	0.0013&	32.00&	0.9673&	0.0062\\
				
				~ &$\times3$ &27.78&	0.9274	&0.0012	&28.55&	0.9368	&0.0052\\
				~ &$\times4$ &25.84&	0.8966&	0.0011&	26.82&	0.9096&	0.0056\\
				
				\hline
				\multirow{3}*{Bicubic} &$\times2$ &31.81&	0.9657	&0.0020	&32.37&	0.9694&	0.0075\\
				
				~ &$\times3$&28.63&	0.9365&	0.0019	&29.07&	0.9413	&0.0065 \\
				
				~ &$\times4$ &26.70&	0.9087&	0.0018	&27.13&	0.9139&	0.0068 \\

				\hline
				\multirow{3}*{Lanczos} &$\times2$ &32.43	&0.9687	&0.0020	&32.74	&0.9706	&0.0074 \\
				
				~ &$\times3$ & 29.05&	0.9402&	0.0018&	29.29&	0.9427&	0.0063\\
				~ &$\times4$ &27.07&	0.9129&	0.0017&	27.33&	0.9157	&0.0067 \\
				
				\hline
			
    \end{tabular}%
		
\end{table*}%

\subsection{Comparison with State of the Art}  
\begin{figure*}[!htbp]
  	
  	\centering
  	\subfigure[Reference/PSNR]{
  		\label{fig:subfig:a} 
  		\includegraphics[width=0.25\textwidth]{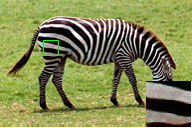}}
  	\hspace{0.5in}
  	\subfigure[GR/25.51 dB]{
  		\label{fig:subfig:a} 
  		\includegraphics[width=0.25\textwidth]{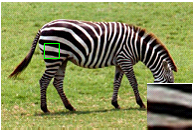}}
  	\hspace{0.5in}
  	\subfigure[Our/25.37 dB]{
  		\label{fig:subfig:c} 
  		\includegraphics[width=0.25\textwidth]{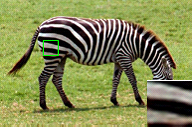}}
  	  \hspace{0.5in}
  	\caption{'Zebra' image from Set14 with upscaling *3}
  	\label{fig:subfig} 
  \end{figure*}

To the best of our knowledge, among the state of the art SISR algorithms that have Matlab source code released, GR \cite{timofte2013anchored} is the fastest. We compare the accuracy and speed of our method with GR. In this experiment, we fix the interpolation function $f$ to be Lanczos. Table \ref{tab:state} shows the comparison, and Fig.\ref{fig:subfig} shows a visual assessment. Relative to GR, the PSNR of our method is generally around $0.2$dB lower for various scaling factors. However, on BSD100 the accuracy performance are very close. This might be attributed to the integration weights learned from BSD200, and suggests the potential for higher performance by using more sophisticated training strategy. It should be noted that the proposed warping model has merely 6 parameters, and the computation is tens of times faster. 
\begin{table*}[!ht]
  \centering
  	\caption{Comparison of PSNR and computation time of GR \cite{timofte2013anchored} the proposed color warping (with the interpolation function fixed to be Lanczos), averaged over four benchmark datasets, for various scaling factor $S$.}
  \label{tab:state}%
  \vspace{1ex}
   \begin{tabular}{|c|c|c|c|c|c|}
   	\hline

   	\multirow{2}*{Dataset}& \multirow{2}*{$S$}&  \multicolumn{2}{|c|}{GR} & \multicolumn{2}{|c|}{Ours}  \\

    \cline{3-6}   
	&  & PSNR & Time & PSNR & Time\\
   	\hline
   	\multirow{4}*{Set5} &$\times2$ &	33.04&	0.34	&32.74&0.0074\\
   	
   	~ &$\times3$ &29.50&	0.22&	29.29&	0.0063\\
   	
   	~ &$\times4$ &	27.49&	0.19&	27.33&	0.0067\\
   	
   	\hline
   	\multirow{4}*{Set14} &$\times2$&	29.85&	0.67&	29.62&	0.0138\\
   	
   	~ &$\times3$&26.92&	0.46&	26.75	&0.0133 \\
  	~ &$\times4$ &25.21	&0.38&	25.08&	0.0124 \\

   	\hline
   	\multirow{4}*{BSD100} &$\times2$ &28.90	&0.45&	28.97&	0.0073 \\
   	
   	~ &$\times3$ &26.38&	0.32&	26.30&	0.0070\\
   
   	~ &$\times4$ &	25.03&	0.26&	24.97&	0.0070 \\
   	
   	\hline
   	\multirow{4}*{Urban100} &$\times2$ &	26.30&	2.31	&26.20&	0.0480\\
   	
   	~ &$\times3$ 	&	23.69	&1.46&	23.50	&0.0427 \\

   	~ &$\times4$ & 22.28 & 1.24  & 22.08 & 0.0428 \\			
   	\hline
   
    \end{tabular}%

\end{table*}%

\section{Conclusion}
\label{sec:conclusion}
In this paper, we have proposed a light weight color image warping technique, which integrates intra-channel interpolation with cross-channel details. Extensive experiments on four benchmark datasets have validated that, the proposed technique substantially improves the most popular image warping algorithms, in very simple form and at trivial computational cost. Our algorithm has also been shown to push the performance of traditional Lanczos upsampling scheme to be comparable with state of the art methods, while being real time. 

The proposed technique can be readily applied to interactive multimedia applications that require both fast image warping and realistic visual effects, such as Computer Gaming and Image Editing. As our integration of cross-channel information is of light weight, it is suitable for multimedia devices with limited computation power. It should be noted that our baseline interpolation, integration formulation and weight training are all very basic. Actually, each step can be extended for higher performance, if the computation cost budget allows. Moreover, our method can also be used as an high quality initialization to many CNN-based methods. 
\bibliographystyle{1}
\bibliography{2}

\end{document}